\pdfoutput=1
\documentclass[journal]{IEEEtran}

\newcommand{\RNum}[1]{\uppercase\expandafter{\romannumeral #1\relax}}
\usepackage{gensymb}

\usepackage{subfigure}
\usepackage{soul}
\ifCLASSINFOpdf
   \usepackage{graphicx}
   \graphicspath{ {images/} }
\else
\fi
\usepackage{amsmath}

\begin{document}

\title{RoboFly: An insect-sized robot with simplified fabrication that is capable of flight, ground, and water surface locomotion}

\author{Yogesh~Chukewad,~\IEEEmembership{Student Member,~IEEE,}
        Johannes~James,~\IEEEmembership{Student Member,~IEEE,}
        Avinash~Singh, 
        and~Sawyer~Fuller,~\IEEEmembership{Member,~IEEE}
\thanks{Authors are with the Department of Mechanical Engineering, University of Washington, Seattle, WA, 98105 USA. (e-mail: yogeshc@uw.edu; jmjames@uw.edu; avinsh@uw.edu; minster@uw.edu)}

\thanks{\textit{Corresponding author: Yogesh Chukewad}}
\thanks{Manuscript received Month XX, 20XX; revised Month XX, 20XX.}}

\markboth{IEEE TRANSACTIONS ON ROBOTICS,~Vol.~xx, No.~x, Month~20xx}%
{Chukewad \MakeLowercase{\textit{et al.}}: Bare Demo of IEEEtran.cls for IEEE Journals}

\IEEEpubid{\copyright~2020 IEEE}

\maketitle

\begin{abstract}
Aerial robots the size of a honeybee ($\sim$100~mg) have advantages over larger robots because of their small size, low mass and low materials cost. Previous iterations have demonstrated controlled flight but were difficult to fabricate because they consisted of many separate parts assembled together. They also were unable to perform locomotion modes besides flight. This paper presents a new design of a 74~mg flapping-wing robot that dramatically reduces the number of parts and simplifies fabrication. It also has a lower center of mass, which allows the robot to additionally land without the need for long legs, even in case of unstable flight. Furthermore, we show that the new design allows for wing-driven ground and air-water interfacial locomotion, improving the versatility of the robot. Forward thrust is generated by increasing the speed of downstroke relative to the upstroke of the flapping wings. This also allows for steering. The ability to land and subsequently move along the ground allows the robot to negotiate extremely confined spaces, underneath obstacles, and to precise locations. We describe the new design in detail and present results demonstrating these capabilities, as well as hovering flight and controlled landing.
\end{abstract}

\begin{IEEEkeywords}
Insect Scale Flapping-Wing Robot, aerial systems: mechanics and control, Ground Locomotion, Air-Water Interfacial Locomotion, Micro-fabrication.
\end{IEEEkeywords}

\IEEEpeerreviewmaketitle

\section{Introduction}

Robots the size of common insects like a honeybee ($\sim$100~mg) have the potential for improved performance relative to larger robots in tasks that benefit from the small size or large deployment numbers. Examples include gas leak detection, assisted agriculture, or an operation around humans without impact hazard. Historically, a key challenge for robots that small was finding a suitable manufacturing method to create the necessary sub-millimeter articulated structure and actuation systems. Additionally, actuators that are in common use in larger-scale robots, such as the electric motors that actuate the propellers in most quad-rotor style drones, do not scale down favorably to insect scale in terms of efficiency or power density \cite{WoodPico2012}. This is because surface area-dependent losses such as Coulomb friction and electrical resistance take on greater importance as scale reduces \cite{TrimmerScaling1989}. Recently, however, a suitable manufacturing process and actuation technology were demonstrated that allowed for controlled flights in an 81~mg robot \cite{ma2013}. This robot was built using a diode-pumped solid-state laser and pin-aligned sheet adhesion to fabricate the necessary components \cite{popUpBook}, and was actuated by piezo-driven flapping wings that emulated the motion of insects~\cite{c1, c5, c6, c7}. The mechanism required to convert the actuator motion to wing motion for generating aerodynamic lift is discussed in \cite{c3}, in which transverse bending of wings as observed in insect flapping is investigated for efficient flapping.

This paper addresses three deficiencies of the basic design introduced in~\cite{ma2013}, introducing a new design that makes robot flies both more versatile and easier to fabricate. 

\subsubsection{Complex fabrication} The insect robot design of \cite{ma2013} suffers from being very difficult to fabricate because it requires hand assembly of a relatively large number of discrete components. It also consists of several failure-prone steps. An alternative was proposed in \cite{popUpBook} that reduced the number of parts by taking inspiration from children's pop-up books. A robotic fly design was demonstrated that consisted of a fabrication step that required actuating a mechanism with only a single degree of freedom. But this design approach is complex, requiring 22 layers with many interdependencies between layers.

\subsubsection{Difficulty in landing} The work in~\cite{ma2013} demonstrating controlled flight by an 81~mg robot relied on feedback control of its upright orientation using retro-reflective marker-based motion capture. When upright, its long axis extends vertically, raising its center of mass and making it challenging to achieve a successful landing without toppling over. Successful landings with that design required leg extensions that nearly doubled the vehicle's size \cite{c20}. An alternative is to use switchable electrostatic adhesion~\cite{graule2016} for perching and takeoffs on vertical or overhanging surfaces, but this adds complexity including a high-voltage source, requires a small amount of additional power to remain attached, and is not required for ground-based landings.

\IEEEpubidadjcol

\subsubsection{Limited mobility autonomy} Mobility autonomy for terrestrial robots can be defined as the ability to traverse unknown and non-smooth terrain~\cite{Bergbreiter}. Here, we define mobility autonomy for insect scale robots as their ability to traverse locomotion with multiple modes which involve aerial, terrestrial and aquatic locomotion. There have been significant developments in small scale robotics, including a 1~g miniature water strider robot~\cite{song2007surface}, a robot that can jump from the surface of water~\cite{koh2015jumping}, a 1.6~g underwater quadrupedal robot~\cite{chen2018controllable}, and a 175~mg flying robot capable of making the transition from water to air~\cite{chen2017biologically}. Here, to avoid actuators contacting water, we focus on the locomotion on the surface of the water.

The work in~\cite{chen2017biologically} demonstrates an aerial-aquatic flight that relies on a sparker while performing the transition from water to air. Though this robot can move itself underwater, it requires de-ionized water to avoid short-circuit and accidentally breaking actuators. Most bodies of water are not this pure, and conduct electricity, causing sudden transient arcs and actuator breakage, requiring a perfect seal on the actuators. Multi-modal locomotive capabilities have been widely studied for biological species such as water striders that rely on surface tension to support their weight. An investigation of the dynamics of water walking creatures is presented in \cite{hu2003hydrodynamics,hu2010hydrodynamics, bush2006walking}. Hydrodynamics of water walking arthropods with characteristic length of the order 1~cm is presented in \cite{hu2003hydrodynamics} and \cite{hu2010hydrodynamics}. Propulsion mechanism in water striders, as presented in \cite{hu2010hydrodynamics}, includes momentum transfer through capillary waves and hemispherical vortices created by the driving legs. Various means of weight support at the water surface, as well as lateral propulsion for various water-walking creatures (not limited insects scale), are discussed in \cite{bush2006walking}. Propulsion mechanisms presented in the work include-- 1) surface slapping (lizards), 2) rowing and walking (most of the water insects), and 3) meniscus climbing (\textit{Pyrrhalta nymphaeae} larvae and \textit{Mesovelia}). A quantitative biomechanical model of insects' interfacial flight is presented in \cite{mukundarajan2016surface}, along with an investigation of water-lily beetles' interfacial flights. It was shown that the interfacial flight is energetically expensive as compared to aerial flights. While above-mentioned research focused on aquatic locomotion of biological creatures, the review by Kwak and Bae in \cite{kwak2018locomotion} connects the biology with robotics by identifying robotic research in aquatic locomotion that can draw inspiration from its biological counterparts. 

\begin{figure}[tb] 
  \centering
  \includegraphics[width=3.2in]{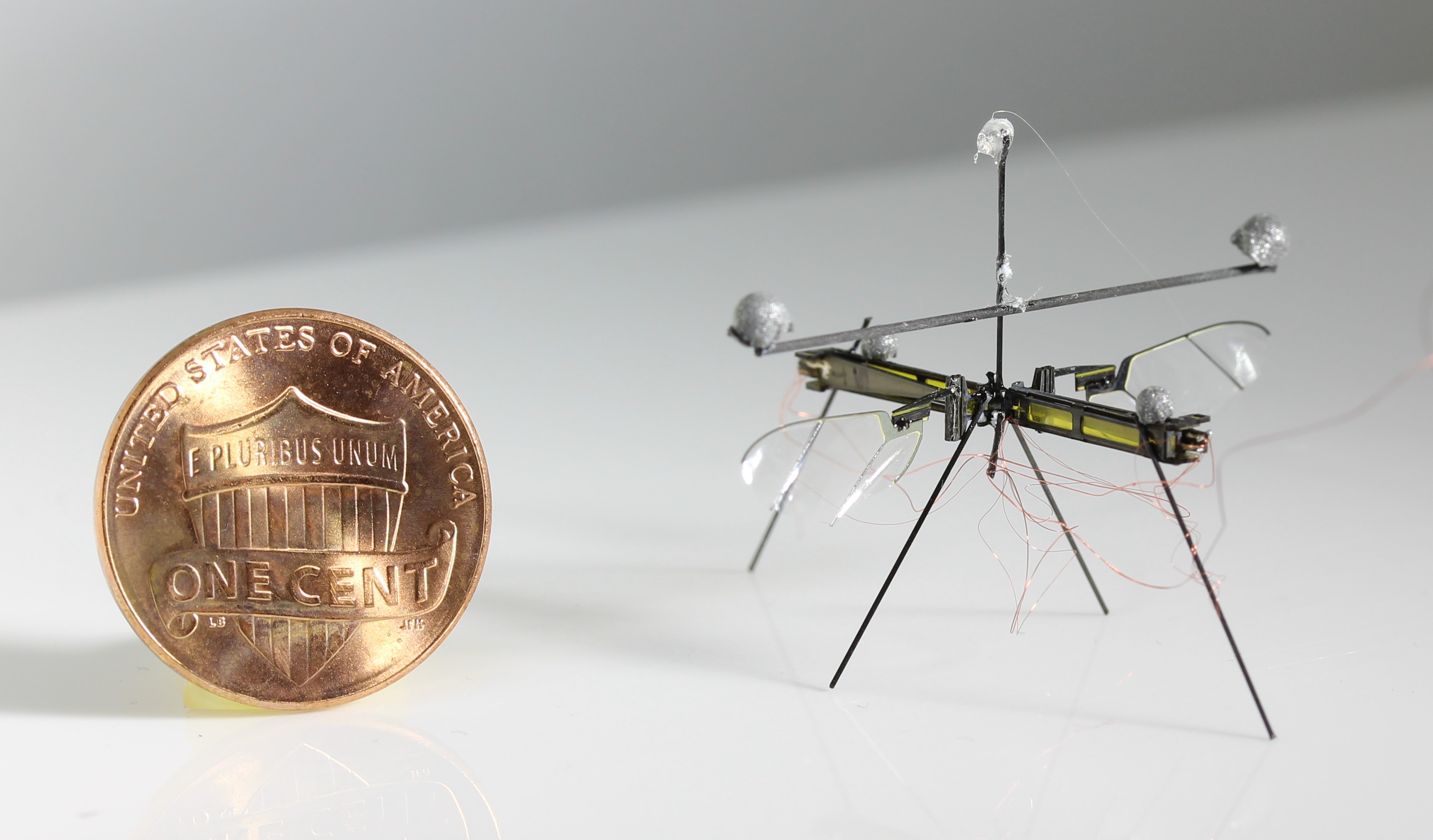}
  \caption{The redesigned system: University of Washington RoboFly. Each wing measures 13~mm in length and is driven by a separate piezoelectric cantilever actuator. By extending the actuators forward and aft, the center of mass is positioned near the base of the wing pair so that there is no net torque during flight. The entire robot weighs 74~mg (without retro-reflective markers). A US penny is shown for scale.}
  \label{robofly}
\end{figure}

In light of the above limitations in fabrication, landing and mobility autonomy in previous designs, this paper describes a new design of an insect-sized flying robot, which we call the University of Washington RoboFly (Fig. \ref{robofly}), that is intended to overcome the deficiencies of previous designs described above. Current paper evolves from authors' earlier conference paper \cite{chukewad_IROS}, which focused on the design and fabrication of the RoboFly, and its capability to perform open-loop landing and ground locomotion. The current submission represents a significant advance over the earlier paper. It, in addition to earlier results, demonstrates how the robot can be modified with a set of passive legs to perform water-air interfacial locomotion. The earlier paper had demonstrated a stabilized takeoff; however, the current paper takes a step further to demonstrate a controlled hovering flight and closed-loop landing.

\begin{figure*}[t]
	\centering
	\includegraphics[width=6.0in]{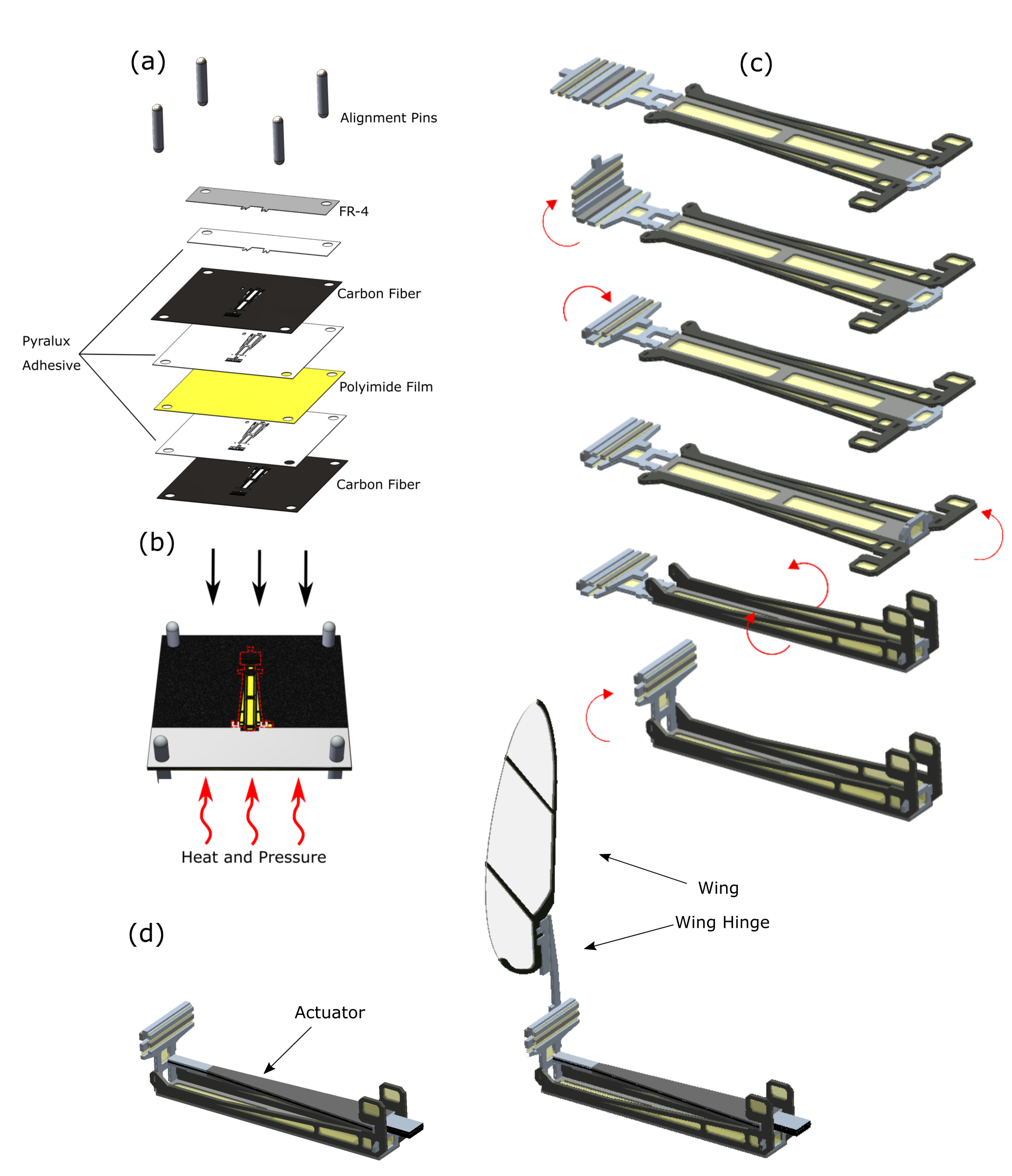}
	\caption{(a) An exploded view of the layup before curing. (b) Layup during the curing process under predetermined pressure and temperature. Release cut to be done on the cured laminate is shown in red dotted lines. (c) The released laminate is shown at the top, followed by the process involving folding of transmission and the airframe. (d) (Left) An actuator is slid into its designated slot on the airframe, (right) wing with its hinge is attached on the transmission.}
	\label{combined_layup_and_folding}
\end{figure*}

Here we report three main contributions to the design of the robot insects, which are embodied in a new design we call RoboFly.

\begin{enumerate}
    \item This design introduces a fabrication process in which the basic wing actuation unit to be composed of a single laminate, simplifying fabrication relative to earlier designs.
    \item This design has a lower center of gravity, which facilitates open-loop landing on the ground.
    \item This robot also possesses better mobility autonomy with its ability to perform multi-modal locomotion which includes aerial, ground and water locomotion. The robot uses its wings to push itself along the ground and water once landed, without additional complexity and weight of a separate walking mechanism.
\end{enumerate}

The rest of the paper is organized as follows. In section \RNum{2}, we introduce the new robot re-design and its simplified fabrication which significantly reduces the number of parts and improves the accuracy of the assembly. In section \RNum{3}, we discuss the ability of the robot to perform multi-modal locomotion, including aerial and ground locomotion. Results from experiments from ground and aerial locomotion are also presented. In section \RNum{4}, we first provide the theory behind small floating objects. We also discuss the design of hydrophobic legs which can be attached to the robot to make it float on the water surface. We also present results from experiments from water locomotion and a transition from aerial to interfacial flight in which the robot lands on the water without breaking the surface tension film. In section \RNum{5}, we discuss the power consumption in the different locomotion modes.

\section{RoboFly Fabrication}

In this section we discuss the basic design of RoboFly as introduced in \cite{chukewad_IROS}. This robot has served as a platform for several studies, including wireless power circuit \cite{JJames_ICRA2018}, a pinhole lens camera \cite{siva_biorob18} on-board. Four half-flies are assembled in the work by \cite{fuller2019four}. In the current study, we focus on its design and consequent expanded locomotion capabilities.

RoboFly consists two identical subunits, each of which consists of a piezoelectric actuator, a carbon fiber airframe, and a wing. A set of four vertical legs are attached for landing and ground locomotion. The process of fabrication and assembly is explained in detail below.

Our design simplifies fabrication by combining the airframe, transmission, and actuator attachment hardware into a single laminate sheet. In the previous design that performed controlled flight~\cite{ma2013}, these consisted many separate parts. Combining these into a single laminate reduces the number of discrete parts and facilitates fabrication during prototyping. Many design features and alignment steps can be built into the design of the laminate. For example, the laminate consists of castellated folds \cite{Sreetharan2012} that impose a precise rotation axis, and mechanical interlocks that can constrain folds to a specific angle. 

\begin{figure*}[t]
	\centering
    \includegraphics[width=5in]{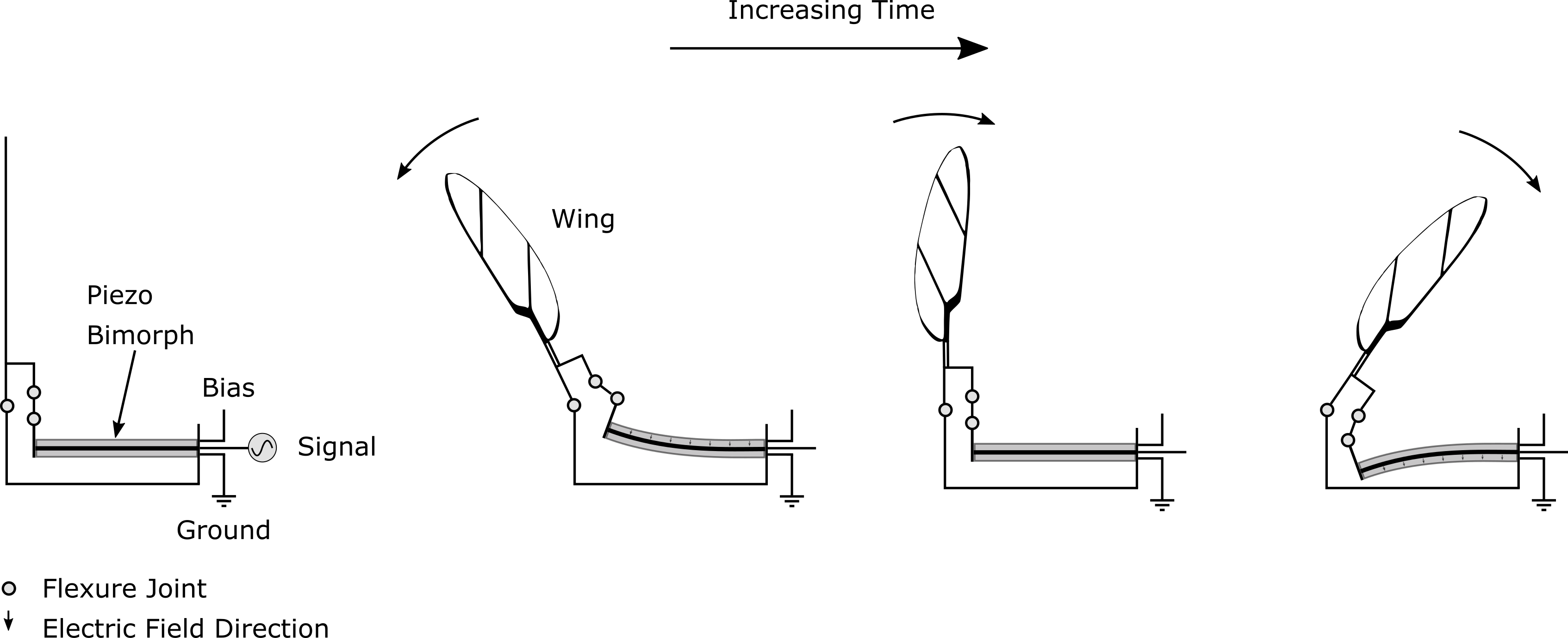}
    \caption{Diagram of the mechanism of piezoelectric cantilever actuation of the wings the design presented in this work. The piezo actuator drives large-amplitude wing motion through small strain changes. The piezo actuator is configured as a bimorph cantilever, consisting of a carbon fiber layer sandwiched between top and bottom piezo sheets. The top surface of the bimorph is charged to a constant high voltage, while the bottom surface is tied to ground  per ``simultaneous drive" configuration. An alternating signal is connected to the middle layer, providing an alternating electric field in the piezo material. This produces alternating small strains through the reverse piezoelectric effect, which is manifested as motion at the tip of the cantilever. A microfabricated transmission amplifies these tip motions into large ($\sim$ 90~deg) wing motions. This diagram shows the mechanism as seen from above; motion of the wings causes airflow downward, into the page}
    \label{motion_actuator}
\end{figure*}

\begin{figure}[bt]
\centering
\includegraphics[height=3.5in]{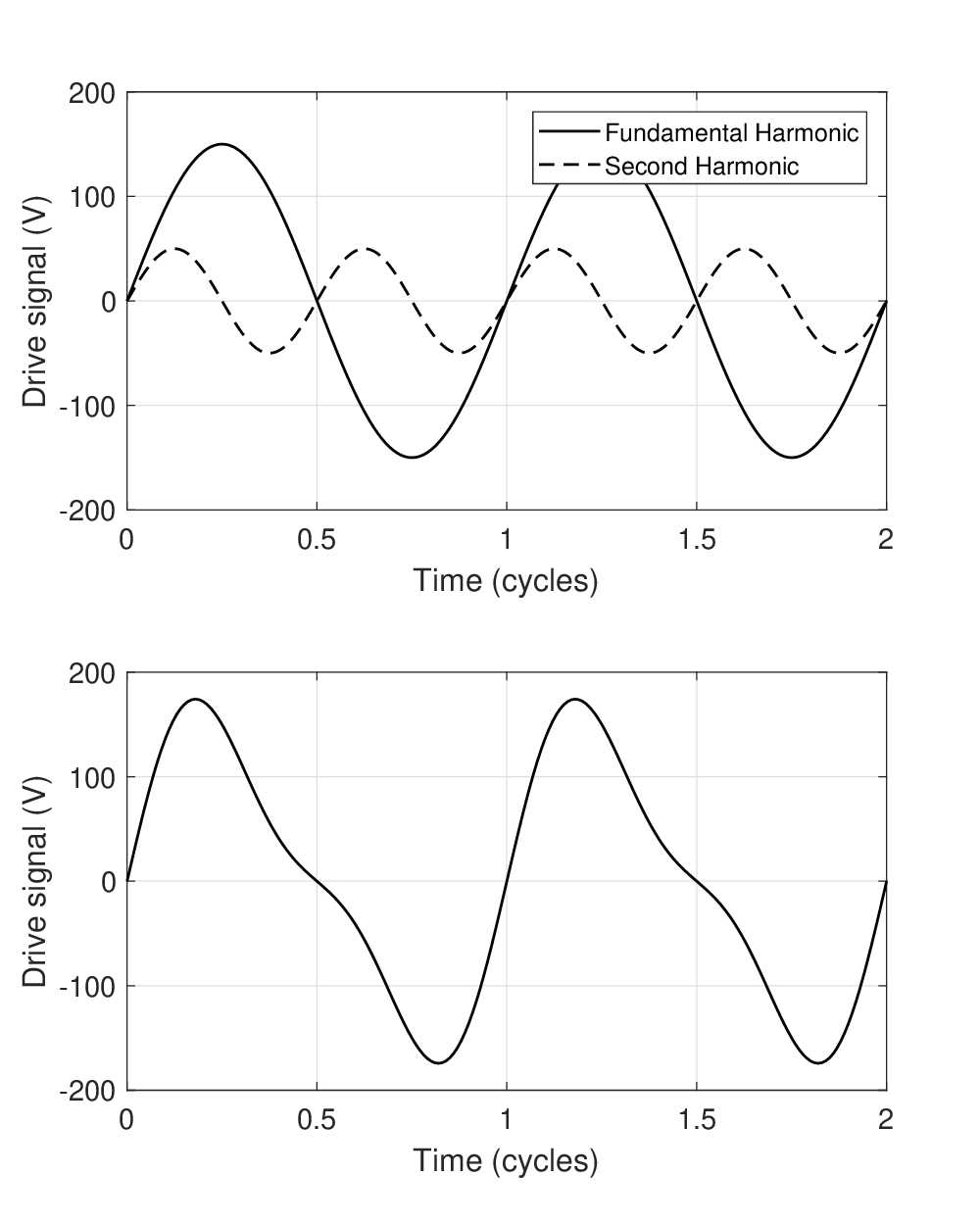}
\caption{The addition of a second harmonic signal causes a differential stroke speed. (top) The sinusoidal drive signal to the wings and the second harmonic at 0.3 times the fundamental amplitude. (bottom) The sum of the two signals.}
\label{signals}
\end{figure}

The laminate is machined and assembled using the following steps:
\begin{enumerate}
  \item Two carbon fiber composite sheets (0\degree-90\degree-0\degree sheets of 27~$\mu$m thick cured prepeg) are laser machined using a diode-pumped solid-state frequency tripled Nd:Yag laser with 355~nm wavelength (PhotoMachining, Inc., Pelham, NH USA). These two sheets constitute the rigid structural material at the top and bottom of the composite layup.
  \item A modified acrylic adhesive (FR1500 Pyralux, DuPont, Inc., Midland, MI) is laser machined with the same pattern as the respective carbon fiber layer features.
  \item A layer of 12.5 $\mu$m polyimide film (Kapton) is laser cut and is placed between the two adhesive layers. The thickness of the Kapton film is chosen according to the flexure feature dimension of the transmission.
  \item Polished stainless steel pins align these layers, ensuring that the features are placed correctly, as shown in Fig.~\ref{combined_layup_and_folding}~(a).
  \item The layup is cured in a heat press at $200\degree$C, $480$~kPa (Fig.~\ref{combined_layup_and_folding}~(b)).
  \item The layup is placed back in the laser system where it is re-aligned rotationally and in translation relative to the beam. Release cuts are machined as necessary. The release cut is shown with red dotted lines in Fig.~\ref{combined_layup_and_folding}~(b).
  \item Each airframe-transmission part is folded by hand with tweezers under a microscope and bonded with cyano-acrylate adhesive (Fig.~\ref{combined_layup_and_folding}~(c)).
  \item An actuator is then carefully placed and bonded to the slots provided on the airframe with extra material to insure a rigid connection at its base (Fig.~\ref{combined_layup_and_folding}~(d)).
  \item A wing is bonded to a wing hinge, and the assembly is then bonded to the transmission (Fig.~\ref{combined_layup_and_folding}~(d) (right)). As in~\cite{wood_2008}, the wing hinge allows the angle of attack to change passively \cite{ma2012}.
  \item Two half-fly assemblies are bonded together at the middle on a specially-designed mating surface.
  \item 30~$\mu$m diameter carbon fiber rods are glued to the static surface of the transmission and at the front and back extremes of the body to form the legs. 
  \item A wire bundle consisting of four 51-gauge insulated copper wire is then carefully soldered onto the actuators' bases to complete the electronic connections. 
\end{enumerate}

\section{Multi-modal locomotion}

\begin{figure}[t]
	\centering
	\includegraphics[height=3.5in]{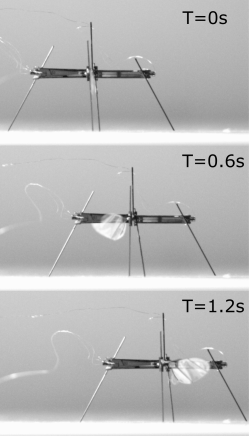}
	\caption{The robot moves forward over the ground when wings are flapped faster in the backward direction than the forward direction. Flapping frequency 60~Hz. In the absence of a steering command, the robot moves in a straight line}
	\label{straight_line}
\end{figure}

\begin{figure}[bt]
	\centering  
	\includegraphics[width=3.2in]{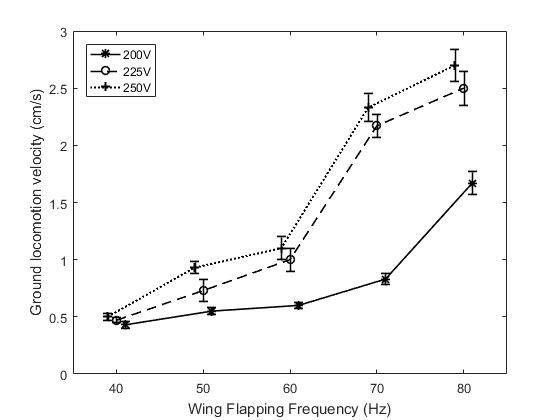}
	\caption{Ground locomotion velocity increases with increasing signal amplitude and flapping frequency. For comparison, liftoff occurs at approximately 140~Hz.}
	\label{speed_image}

\end{figure}

\begin{figure}[bt]
    \centering
    \includegraphics[width=2.5in]{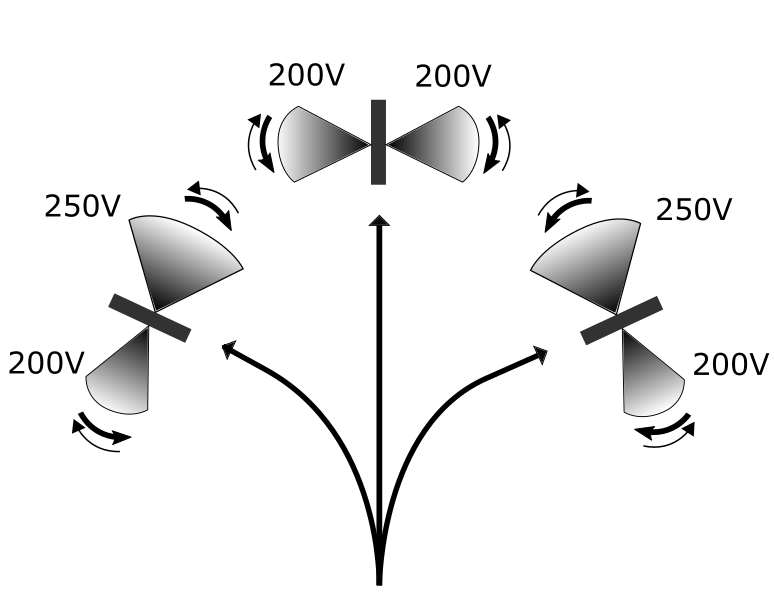}
\caption{A top view of ground locomotion and steering. Steering can be performed by driving the wings with unequal signals. Thickness of the arrows corresponds to the stroke speed. Here, the rearward stroke is faster than the forward stroke, causing forward motion.}
\label{steering}
\end{figure}

\begin{figure*}[bt]
	\centering
	\includegraphics[width=6.5in]{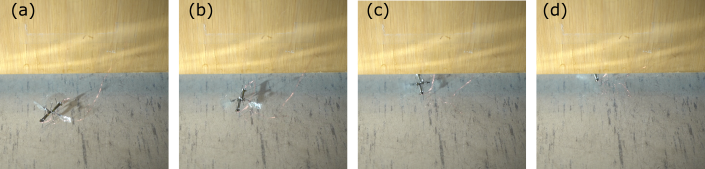}
	\caption{Ground locomotion allows the robot to navigate under aerial obstacles.
    The robot is shown ambulating under a closed door, which would not be possible by flying.}
	\label{aerial_obstacles}
\end{figure*}

\begin{figure*}[bt]
	\centering
	\includegraphics[width=6.5in]{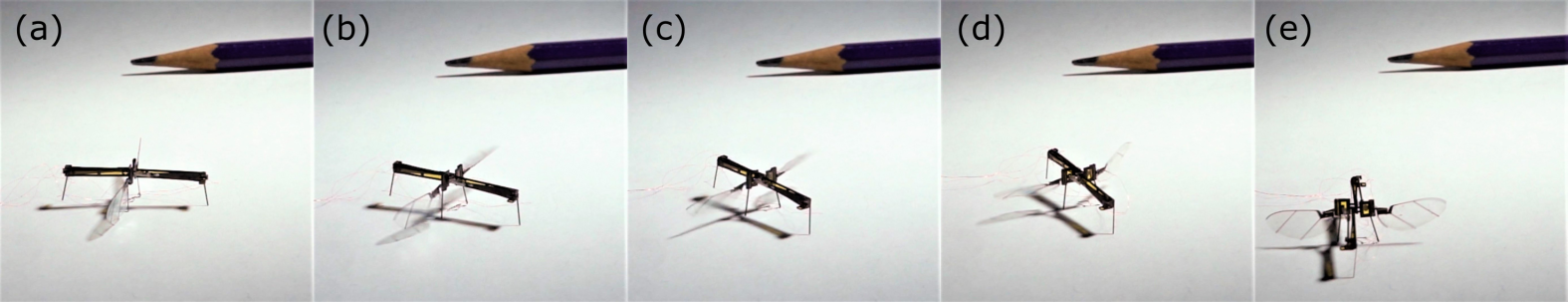}
    \caption{Robot turns right by 90\degree. A pencil tip is shown in the background for scale.}
    \label{animals2}
\end{figure*}

This section discusses experimental apparatus and results. First, we describe the operation of the robot and the hardware involved in the experiments, followed by experimental results of the robot performing different types of multi-modal locomotion. 

\subsection{Operation}

The RoboFly has a set of four passive legs in vertical plane as shown in Fig.~\ref{robofly}. This wide stance provides fabrication simplicity and facilitates landing. 

The piezo actuators were driven by a desktop computer equipped with a digital-to-analog conversion board (NI 6259) running Simulink Real-Time (MathWorks, Natick, MA, USA) and amplified using three high voltage amplifiers (Trek 2205, Lockport, New York). One amplifier supplies the DC `bias' signal to both actuators; the other two amplifiers each supply the separate sinusoidal drive signals to the two wings. 

Ground locomotion is performed by flapping the wings at a lower frequency than is needed for takeoff. We chose the stroke amplitude for ground locomotion to be same as that for flight. Stroke amplitude is determined by the amplitude of the drive signal voltage (Figs. \ref{motion_actuator}, \ref{signals}).

The wing flapping frequency is varied depending on the mode of the locomotion that we want to carry out. However, it is kept constant for a particular mode which can be one of the following set of actions at any time instance-- ground locomotion, water locomotion, and flight. Maneuvers while performing these actions were carried out by varying the voltages at which the actuators are driven. All actions besides flight are performed at non-resonant frequency to avoid accidental lift-off. Each actuator is driven at a voltage signal, $V(t)=V_0+A_0\sin(\omega t)+A_1\sin(2\omega t)$, where $V_0$ is the offset voltage, $A_0$ the amplitude, $A_1=\mu A_0$ the amplitude for the second harmonic term; a typical value $\mu$ ranges from $-$0.3 to 0.3. By adding a second harmonic at double the frequency, so that either the downstroke or upstroke is faster (Fig.~\ref{signals}), the robot is driven forward or backward as result of aerodynamic drag on the wings. For example, forward motion occurs when the signal to the wings drives them rapidly backwards. A similar mechanism was proposed to induce torques about a vertical or yaw axis in \cite{ma2013} and \cite{yaw_authority_wood_2019}.

In case of aerial locomotion, the wings are flapped at resonant frequency to generate maximum lift. One way to determine the resonant frequency that maximizes the lift for flapping-wing micro aerial vehicles (FWMAV) is presented in \cite{finio2011system}, in which the authors performed the system identification of an FWMAV to fit a linearized second order model.

We hypothesized that forward motion on the ground is due to the robot momentarily exceeding coulomb friction during the fast period of the wing stroke. To determine whether this motion was primarily driven by inertial or by aerodynamic forces, we performed an experiment in which the wings were replaced by carbon rods with identical mass and moment of inertia. When supplied with driving signals that moved the robot when it was equipped with wings, it was observed that the robot with carbon rods did not move significantly from its initial position. This indicates that the forces causing the ground locomotion are mainly due to the aerodynamic drag acting on the wings. To understand why this should be so, we note that the Reynolds' number of the wing is approximately 3000 \cite{wood_2008}, that is, dominated by inertial forces. This indicates that drag is proportional to the square of the wing velocity according to $f_d=\frac{1}{2}C_D\rho A v^2$, where $C_D$ is the aerodynamic drag coefficient, $\rho$ is the air density, $A$ is the frontal area of the wing, and $v$ is the velocity of the wing. Therefore, a faster wingstroke with a higher $v$ will produce higher drag than a slow stroke.

\subsection{Ground Locomotion}

Fig.~\ref{straight_line} and the supplementary video~\cite{video} show the robot performing ground locomotion along a straight line. In these trials, the wings were flapped at 60 Hz with an amplitude 210~V. Fig.~\ref{aerial_obstacles} and the video~\cite{video} show that ground ambulation allows the robot to navigate under a closed door.

To determine how the driving signal affects locomotion, the RoboFly was driven in the forward direction with a range of different voltages and frequencies. Displacements were measured with a ruler, and the speed was calculated by dividing by the time taken. The results show that robot velocity increases with increasing flapping frequency and amplitude (Fig.~\ref{speed_image}). We conjecture that the large velocity increases that occur at different amplitudes are the result of the robot overcoming coulomb friction at a critical phase of wing flapping. The small increase from 225~V to 250~V is likely attributable to the small resulting additional stroke amplitude. 

Steering is performed by varying the signals given to each flapping wing independently. To steer the body to the left, the left wing is flapped at a reduced drive signal amplitude relative to the right wing (Fig.~\ref{steering}). The rate of rotation is determined by the relative drive signal amplitude difference in the two wings. A sharp turn can be achieved by keeping one wing stationary while the other wing flaps. The extreme continuation of this would be rotation about a vertical axis passing through the center of the body, for which the wings are flapped 180\degree out of phase. A continuous range of turn angles can be achieved by modulating the difference between left and right wing drive signals.

Fig.~\ref{animals2} and the supplementary video~\cite{video} show that the robot is able to steer in addition to moving forward. Here, the wings were flapped at 70 Hz as above, but the left wing was flapped with larger drive signal amplitude (250~V) whereas the right wing was flapped at a lower value (200~V). Similarly, the robot was observed to steer in the other direction when these amplitudes were reversed. Moving backwards was also achieved using the appropriate driving signal.

\begin{figure*}[ht]
	\centering
	\includegraphics[width=6 in]{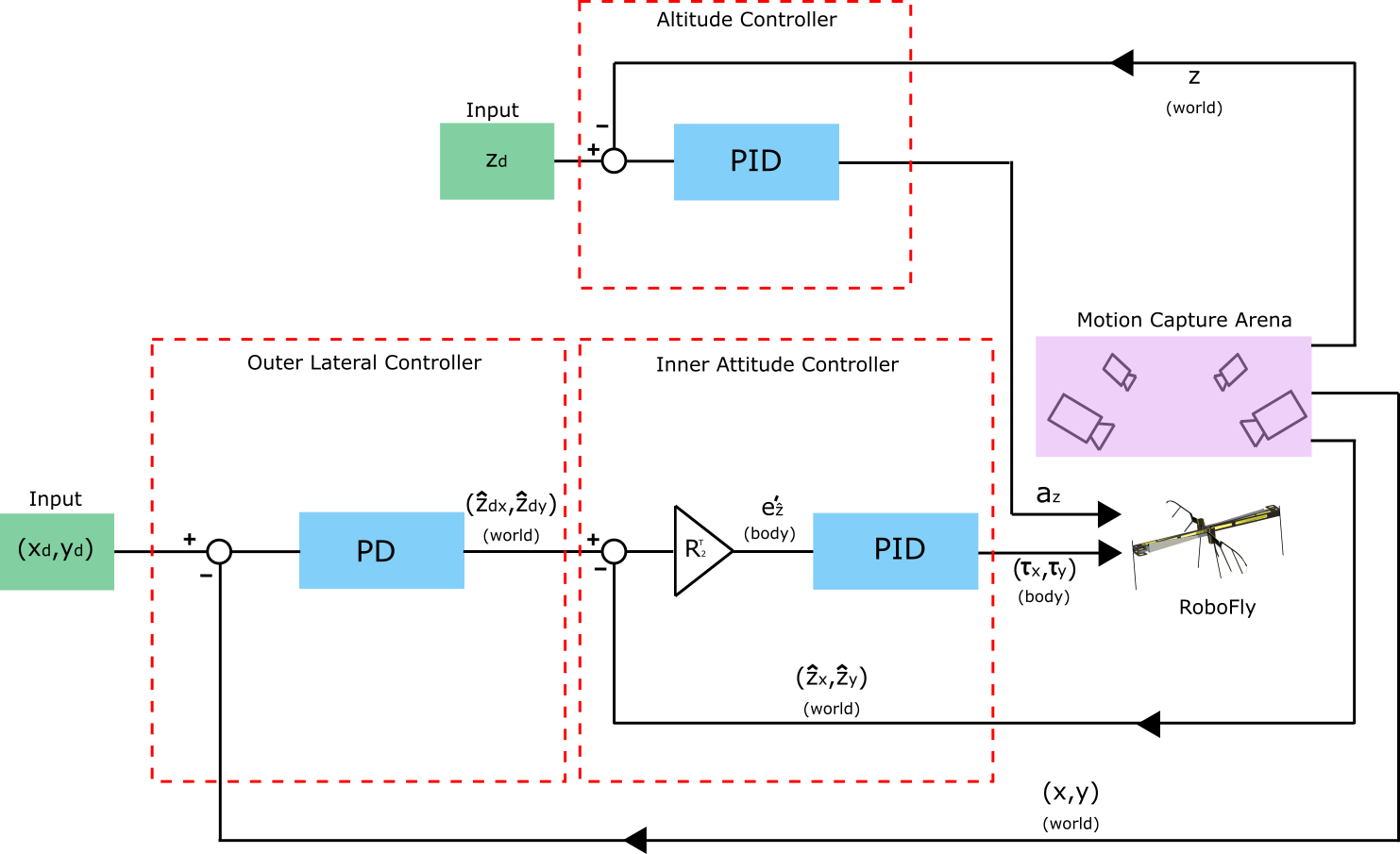}
	\caption{Controller used for hovering. Desired position $(x_d, y_d, z_d)$ is fed as the input to the controller. Altitude controller, as shown at the top, achieves desired altitude $z_d$ by generating vertical acceleration $a_z$. The lateral position controller works on the desired lateral coordinates to compute desired thrust vector orientation, which is fed into the inner attitude controller which determines roll and pitch torques $(\tau_x, \tau_y)$. Motion capture system which tracks the position and orientation of the robot is used for the feedback.}
	\label{control_block}
\end{figure*}

\subsection{Takeoff and Flight}
Hovering at a specified location in space was performed using feedback from a motion capture (MoCap) system (four Prime 13 cameras, OptiTrak, Inc., Salem, OR) which tracks retroreflective markers attached on the robot. This MoCap system sends position and orientation information over Ethernet at 240 Hz to the host desktop computer which runs Simulink Real-Time.

RoboFly is an under-actuated system. However, it can move to any point in space by changing its attitude and tilting the thrust vector in the desired direction of motion. For altitude control the robot uses a  proportional-integral-derivative (PID) controller to achieve desired height. The controller used by the robot for position control is shown in the block diagram in Fig. \ref{control_block}. The MoCap system, through the host computer, continuously sends the position and orientation data over the ethernet to the Simulink Target computer which runs the controller. When the controller is fed with the reference point $[x_d(t), y_d(t), z_d(t)]^T$, it varies the signal amplitude to react to the error in altitude, as mentioned above. At the same time, the controller varies the wing signal according to two control loops--- inner attitude loop and outer lateral position loop, as shown in the block diagram.

As for the altitude controller which runs in parallel with the attitude controller, the height of the robot is controlled using a PID controller. The thrust generated by the wings of the robot is approximately linear with the wing amplitude. Vertical acceleration can be written by the control law as follows:

\begin{equation}
    a_z = k_{ph}e_z + k_{dh} \Dot{e}_z + k_{ih}\int_{0}^{t} e_z dt
    \label{eq:altitude_control_law}
\end{equation}

where $e_z  = z_d - z$ is the error between desired height $z_d$ and current height $z$; $k_{ph}$, $k_{dh}$ and $k_{ih}$ are proportional, derivative and integral gains, respectively. $z$ values received from the MoCap system are first filtered using a low-pass Butterworth filter before taking derivatives. 

As for the cascaded lateral and attitude controllers, the outer loop receives the current and desired lateral positions in the world coordinate system. The error goes through a proportional-derivative (PD) controller, which generates a desired change in attitude vector trajectory $\mathbf{\hat{z}_d(t)}$ which is fed into the inner attitude control loop. The outer loop assumes that the inner loop responds to attitude changes almost instantaneously.

In the outer loop, the lateral position of the robot is controlled by determining the desired inclination trajectory $\mathbf{\hat{z}_d} = [\hat{z}_{dx} \hspace{0.05in} \hat{z}_{dy}]^T $. This is performed by a PD controller in the world coordinate system:

\begin{gather}
\mathbf{\hat{z}_d} =
 \begin{bmatrix} \hat{z}_{dx} \\ \hat{z}_{dy}  \end{bmatrix} =
 k_{pl} \begin{bmatrix} x_d - x \\ y_d - y  \end{bmatrix} +
 k_{dl} \begin{bmatrix} \dot{x}_d - \dot{x} \\ \dot{y}_d - \dot{y}  \end{bmatrix},
\end{gather} where $(x,y)$ and $(x_d, y_d)$ are the current and desired lateral positions in global coordinate system, respectively; $k_{pl}$ and $k_{dl}$ are the proportional and derivative gains, respectively.

In the inner loop, attitude is controlled by rotating the robot's thrust vector towards the desired lateral position. In other words, objective of the inner loop is to align the thrust vector $\mathbf{\hat{z}} = [\hat{z}_{x} \hspace{0.05in} \hat{z}_{y}]^T $ along the desired inclination trajectory $\mathbf{\hat{z}_d}$. So far we have everything in the world coordinate system; however, desired roll and pitch torques need to be determined in body-attached frame.

The robot's attitude is parameterized by a rotation matrix $\mathbf{R}$
that relates body and world coordinates according to $\boldsymbol{v}=\mathbf{R}\boldsymbol{v}'$,
where we define $\boldsymbol{v}$ to be any vector expressed in world
coordinates, and $\boldsymbol{v}'$ is the same vector expressed in
body-attached coordinates. The matrix $\mathbf{R}$ and body angular
velocity $\boldsymbol{\omega}'$ were computed from the quaternion
representation provided by the motion capture system. The error ${e}'_{\hat{\mathbf{z}}}$ between desired trajectory and the current thrust vector position, in  body-attached frame is determined as follows:

\begin{equation}
\boldsymbol{e}'_{\hat{\mathbf{z}}}=\begin{bmatrix}e'_{\hat{z}x}\\
e'_{\hat{z}y}
\end{bmatrix}=\mathbf{R}_{2}^{T}\left(\begin{bmatrix}\hat{z}_{dx}\\
\hat{z}_{dy}
\end{bmatrix}-\begin{bmatrix}\hat{z}_{x}\\
\hat{z}_{y}
\end{bmatrix}\right),
\label{eq:world_to_body}
\end{equation}
where $\mathbf{R}_{2}$ is the upper-left 2$\times$2 block of the rotation matrix $\mathbf{R}$.

The inner faster loop for determining the roll and pitch torques in body-attached frame works as follows:

\begin{equation}
\begin{bmatrix}\tau_{x}\\
\tau_{y}
\end{bmatrix}=k_{pa}\begin{bmatrix}-e'_{\hat{z}y}\\
e'_{\hat{z}x}
\end{bmatrix}-k_{da}\begin{bmatrix}\omega'_{x}\\
\omega'_{y}
\end{bmatrix}+k_{ia}\int_{0}^{t}\begin{bmatrix}-e'_{\hat{z}y}\\
e'_{\hat{z}x}
\end{bmatrix}dt, \label{eq:attitude}
\end{equation} where $k_{pa}$, $k_{da}$ and $k_{ia}$ are the proportional, derivative and integral gains, respectively.  Note that in this controller, \emph{torque} about the $x$-axis depends on inclination along the $y$-axis and vice versa. The attitude error is the difference between desired and actual inclination. If the robot is facing in the world \emph{x}-direction, the \emph{y}-direction will be towards the left of the robot. In a situation where the lateral reference point in on the right of the robot, the attitude error vector will have a component in \emph{--y} direction. It will require the robot to perform positive roll about body \emph{x} axis. This justifies the negative sign with $e'_{\hat{z}y}$ in the above expression.

The controller described above produces a thrust $a_z$ and torques $\tau_x$ and $\tau_y$ that are normalized by body mass and moment of inertia, respectively. To map these accelerations into voltage values supplied to the piezoelectric actuators. We assume that the thrust force is linearly proportional to amplitude. To estimate the moment arm for computing torques, we approximately determined the distance between the center of mass and the aerodynamic center of thrust for individual wing, using CAD model of the robot. The yaw motion of the robot is left uncontrolled in these experiments, as it does not affect the lateral and vertical control of the robot.

\begin{figure}[t]
	\centering
	\includegraphics[width=2.5in]{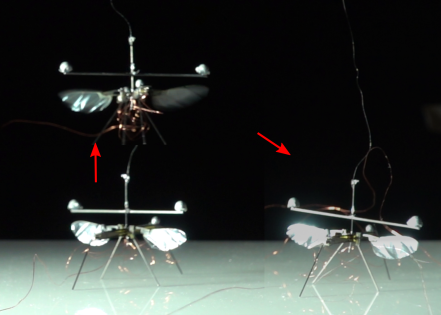}
	\caption{A demonstration of closed-loop takeoff and landing}
	\label{landing_closedloop}
\end{figure}

\begin{figure}[bt]
	\centering
	\includegraphics[height=3in]{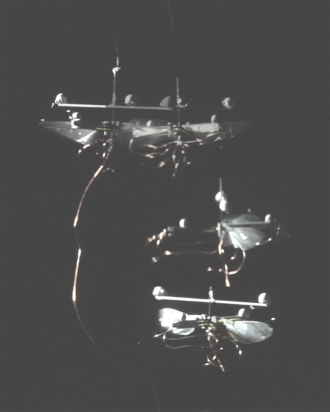}
	\caption{Flight in which the robot takes off and hovers about a reference point in space with the help of feedback from a motion capture arena. The robot is subject to a small yaw bias torque that caused it to rotate leftward in this video \cite{robofly_hovering}.}
	\label{hovering_composite}
\end{figure}

\begin{figure*}[bt]
    \begin{subfigure}
        \centering
        \includegraphics[width = 2.3in]{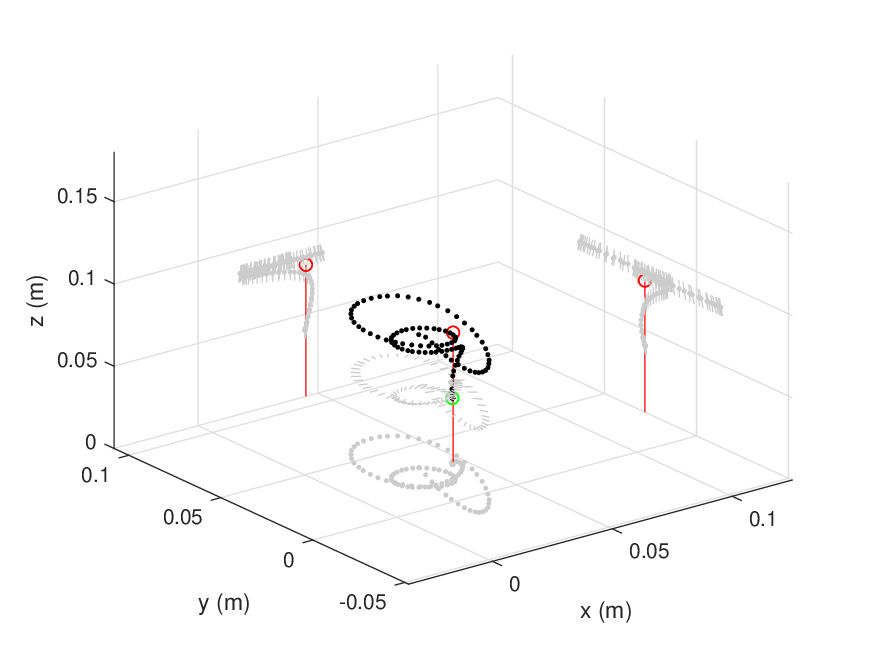}
    \end{subfigure}
    \begin{subfigure}
        \centering
        \includegraphics[width = 2.3in]{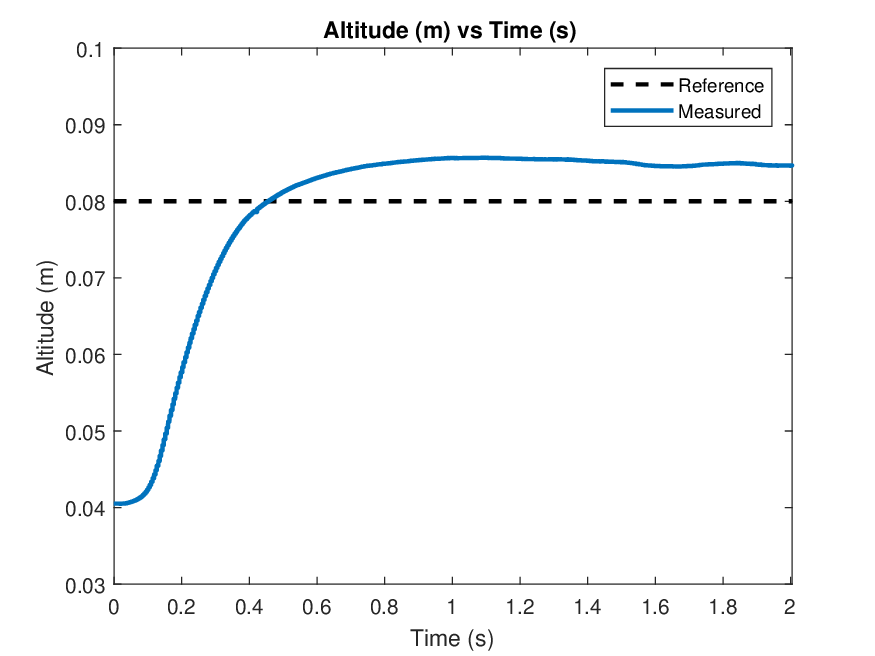}
    \end{subfigure}
    \begin{subfigure}
        \centering
        \includegraphics[width = 2.3in]{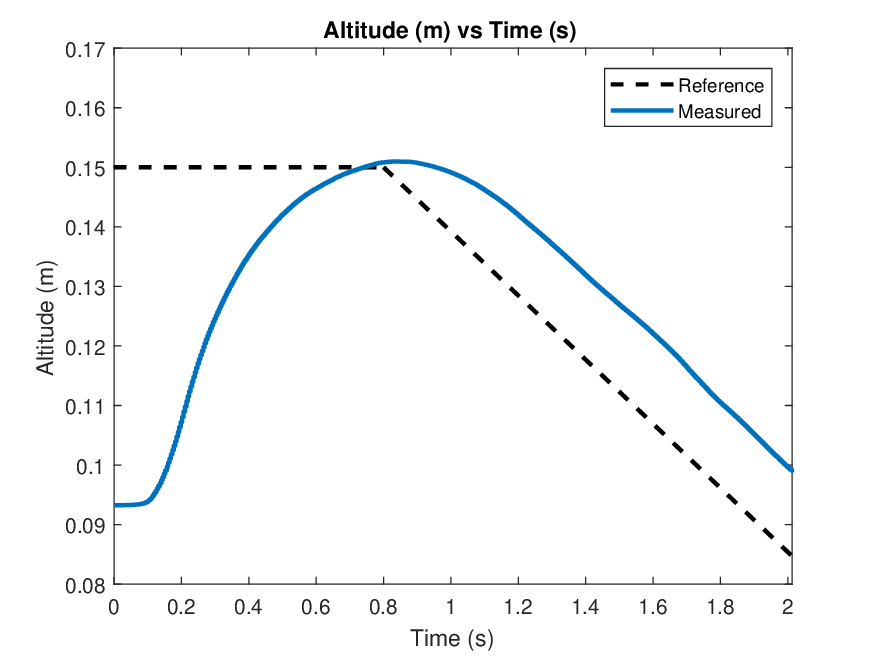}
    \end{subfigure} 
    \caption{(a) Trajectory plot of RoboFly taking off and hovering about an aerial reference point. This plot corresponds to the video from which the frames in Fig.~\ref{hovering_composite} are taken. (b) Measured and reference altitudes vs time from the hovering experiment discussed above. (c) Measured and reference altitudes vs time from the hovering experiment where the altitude is dropped linearly after some time to demonstrate the control over the altitude, which is essential for a controlled landing}
    \label{hovering_plots}
\end{figure*}

We demonstrated controlled takeoff in Fig.~\ref{landing_closedloop}. Frames from a hovering flight are shown in Fig.~\ref{hovering_composite} where the robot was flown for around 2 seconds. The 3-dimensional trajectory plot of this flight is shown in Fig.~\ref{hovering_plots}~(a). The robot was commanded to fly at a height of $4$~cm from the starting point. It can be seen in Fig.~\ref{hovering_plots}~(b) that the robot successfully maintained the height at approximately $4.5$~cm from the takeoff point. The RMS position errors during the last one second of the flight were 1.8, 1.75, and 0.5 cm for its x, y, and z positions, respectively.

\begin{figure}[t]
	\centering
	\includegraphics[height=3in]{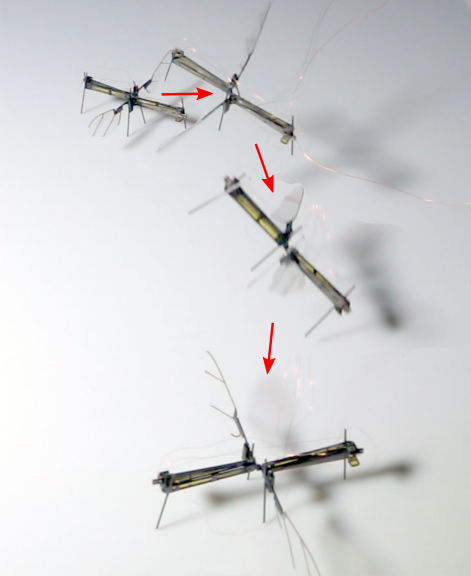}
	\caption{A demonstration of open-loop takeoff and landing}
	\label{landing_openloop}
\end{figure}

\begin{table}[t]
\renewcommand{\arraystretch}{1.3}
\caption{Comparison of number of layers and discrete parts required in different construction methods for creating insect-sized flying robots}
\label{table_comparison}
\centering
\begin{tabular}{c c c c}
\hline
\hline
 & Ma \cite{ma2013} & Sreetharan \cite{Sreetharan2012} & This work\\
\hline
Number of Layers: & 5 & 22 & 7\\
Number of Distinct Parts: & 14 & 1 & 8 \\
\hline
\end{tabular}
\end{table}

\subsection{Landing}

One of the objectives of the re-design reported in this work was to give the robot the capability of landing upright even in the event of loss of control, without the need of long extended-out legs as in \cite{pakpong2013adaptive_landing}. An upright landing allows for an easy transition to the next desired task, such as walking, sensing, or subsequent flights. This is facilitated by our robot's low center of mass, which makes it harder for the robot to topple. 

We demonstrated an uncontrolled takeoff as shown in Fig.~\ref{landing_openloop}. In this video, the robot is seen to be flying with unstable attitudes and landing shortly after. Under feedback control, the robot remains approximately level as shown in Fig.~\ref{landing_closedloop}. Under these conditions, the robot was able to perform a landing by gradually lowering the commanded altitude as shown in Fig.~\ref{hovering_plots}~(c) which was plotted for another experiment where the robot was flown for a longer duration. 

Additionally, we showed that the robot was able to land even when feedback control was not present, indicating that the lowered center of gravity of our design improves landing robustness (Fig.~\ref{landing_openloop}).

\begin{figure*}
    \centering
    \includegraphics[width=4in]{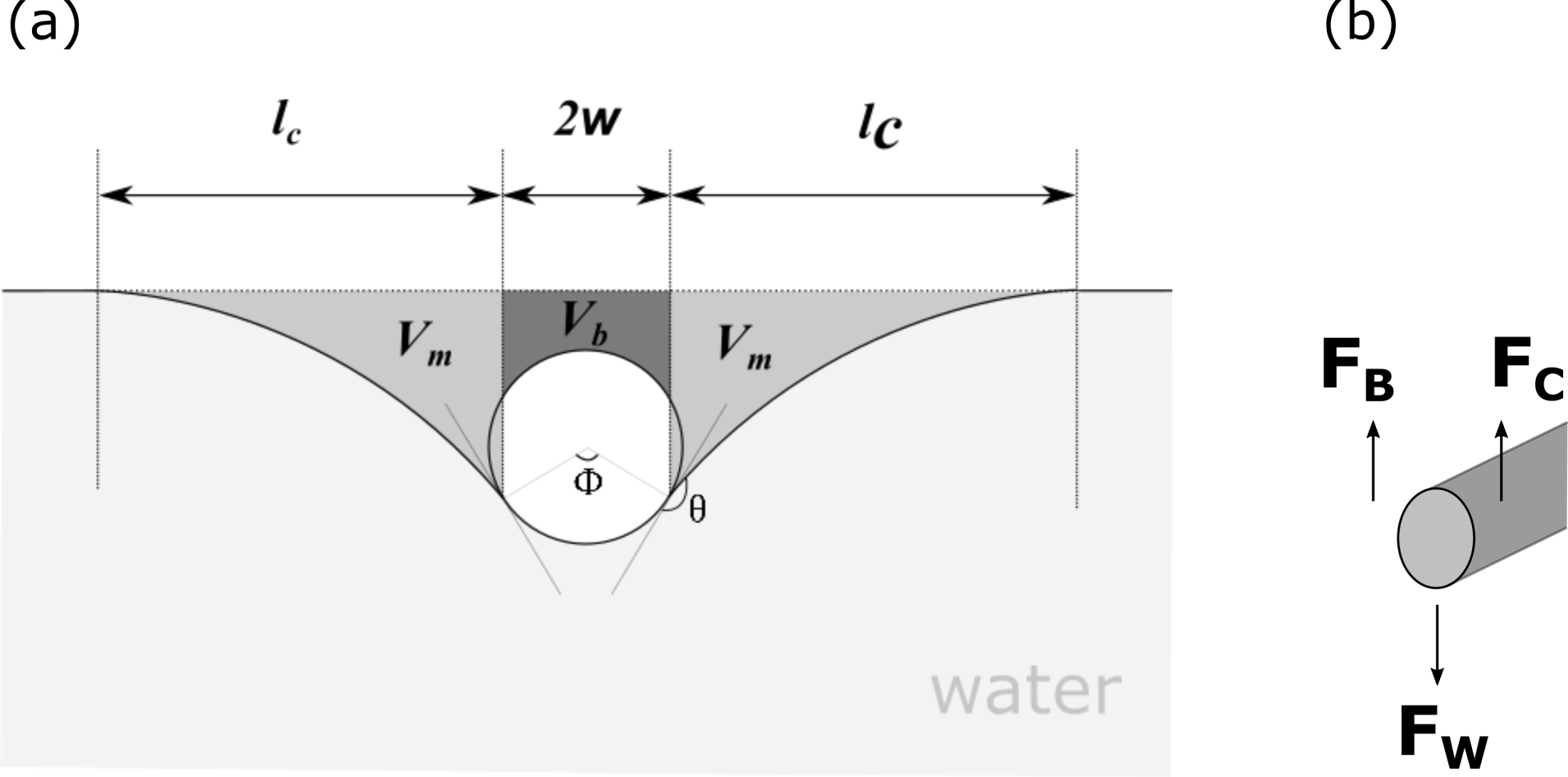}
    \caption{(a) Schematic showing a static state of the cross sectional view of a horizontal leg of water-walking arthropod. $w$ is the radius of the leg, $l_c$ is the capillary length, $\theta$ is the contact angle, $\phi$ is the submerged angle, $V_b$ is the water volume displaced inside the contact line and above the body, and $V_m$ is the water volume displaced  outside the contact line. (b) Vertical loads on the supporting legs. Here, $F_C$ is the curvature force due to surface tension, $F_B$ the buoyancy force, and $F_W$ the weight distribution at the point of contact.}
    \label{stat_dyna}
\end{figure*}

\section{Locomotion on the Surface of Water}

In this section, we show that by adding legs of the appropriate size and shape, the robot can gain an additional locomotion capability: air-water ambulation along the surface of the water.

\subsection{Biological waterstriders}

Our design is inspired by insects, such as stoneflies and mayflies, that use their flapping wings to push themselves along the air-water interface \cite{mukundarajan2016surface}. Like these animals, our robot moves along the surface by propelling itself with its flapping wings. Compared to an airborne RoboFly~\cite{chukewad_IROS} which has six spatial degrees of freedom, an interfacially flying RoboFly has only three degrees of freedom because it is constrained to move along a surface. However, the surface tension forces acting on the legs of the RoboFly still makes it difficult to model the locomotion. Mukundarajan et al. \cite{mukundarajan2016surface} present a dynamic model for an interfacial flight for actual biological insects. According to this model, forces acting on an actual biological insect performing interfacial flight are as follows: 1) horizontal air drag acting on the wings, 2) horizontal capillary-gravity wave drag, 3) water drag acting on the legs at the contact with the water surface in the opposite direction to that of the motion, 4) body weight in vertical direction, 5) vertical resultant force due to surface tension, and 6) vertical water drag. The static case of the RoboFly locomotion is presented in this section.

In case of an object not moving or oscillating vertically, it can float on the water surface if its weight is balanced by the sum of two types of forces exerted on it. The first of these forces is buoyancy force. According to Archimedes' principle, the buoyancy force exerted on an object is equal to the weight of the water displaced. Therefore, objects with density lower than that of water tend to float, and those with density larger than that of water sink. Water-walking arthropods have density larger than water, and therefore they would sink unless supported by the second type of force which is due to the surface tension of water. This phenomenon is explained in detail in the following subsection on legs design.

\begin{figure}[t] 
  \centering
  \includegraphics[width= 3.3 in]{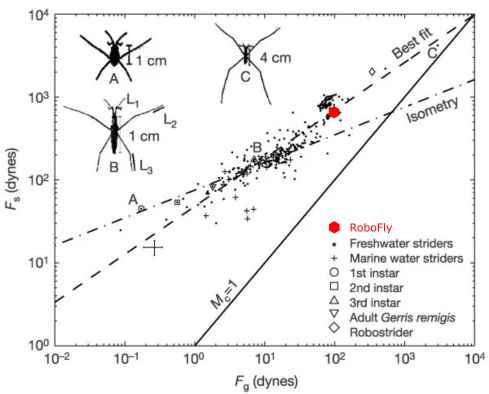}
  \caption{The relation between maximum curvature force and body weight for 342 species of water striders as reported in \cite{hu2003hydrodynamics}. (Figure reproduced with permission from John Bush, MIT). RoboFly is shown in the red.}
  \label{Hu_Bush_plot}
\end{figure}

\begin{figure}[t] 
  \centering
  \includegraphics[width= 2.3 in]{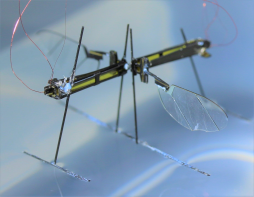}
  \caption{ The RoboFly design weighing $\sim$95 mg and capable of performing multi-modal locomotion including aerial, ground and air-water interfacial flights. Each wing is driven by a separate piezoelectric bimorph actuator. The surface tension force at the horizontal legs at the bottom supports the weight of the robot.}
  \label{Fly_water}
\end{figure}

\begin{figure}[tb]
    \centering
        \includegraphics[width=2.3 in]{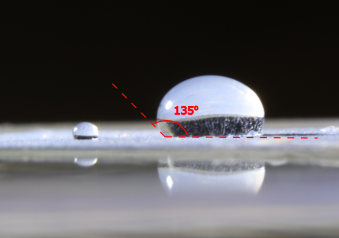}
        \caption{Contact angle measured and shown on a steel shim coated with hydrophobic spray \textit{Rust-Oleum NeverWet}}
        \label{contact_angle}
\end{figure}

\subsection{Supporting legs design using static analysis}

Legs play an important role in keeping the RoboFly on the water surface without breaking the surface tension film. Rigid and compliant legs were considered in \cite{song2007surface}. However, that study was conducted on a robot which used actuating legs for propulsion. Here, we explore the use of wings for propulsion. Therefore, to keep the design simple, a rigid and passive set of three horizontal legs are attached to the RoboFly. These legs lie in the same plane at its bottom so that every part of the legs is in contact with the ground when placed on a flat surface. Cylindrical rods are chosen over square cross-section ones for the legs to avoid complication arising due to a potential case where the robot rests or lands on an edge of a leg. A simple configuration of three parallel legs is considered for this study.

Once we know the configuration and shape of the legs, next design parameters to be considered are 1) material, 2) diameter, 3) length, and 4) distance between the rods. 

The diameter of the legs plays an important role in determining the ratio of buoyancy force and curvature force. The capillary length $l_c$, as shown in Fig. \ref{stat_dyna}~(a) is independent of the leg diameter and contact angle. To keep the surface unbroken it is important to have the curvature force significantly larger than the buoyancy force. A carbon fiber rod of diameter 0.5~mm was chosen for the legs. This diameter corresponds to a Bond number, $Bo\approx0.31$, as calculated later in this section. Legs made out of stainless steel were used in \cite{song2007surface}. Here, carbon fiber rod was chosen for its strength-to-weight ratio: it weighs about 3.6~mg/cm, whereas a stainless steel rod of the same diameter weighs about 15~mg/cm. Since the goal of this robot is also to perform multi-modal locomotion which includes flying, it is important to minimize mass.

While determining the length of legs, it is important to understand the forces acting on the legs while resting on the water surface. Surface tension causing curvature force is assumed to be the primary source of support to balance the weight. Buoyancy forces are not significant for the floating bodies of sub-gram weight, which is also demonstrated in \cite{chen2018controllable}. Buoyancy force depends on the volume of water displaced because of the floating body, it corresponds to volume $V_b$ as can be seen in Fig. \ref{stat_dyna}~(a). Curvature force, on the other hand, corresponds to volume $V_m$ displaced outside the contact line. This volume $V_m$ depends on the capillary length, $l_c$, which is determined as $l_c = (\sigma / \rho g)^{0.5} \approx 2.6$~mm for objects floating on water surfaces, where $\sigma$ and $\rho$ are coefficient of surface tension and density of water, respectively. It can be seen from Fig.~\ref{stat_dyna}~(a) that $F_B / F_c \sim V_b / V_m \sim w/l_c = 0.25/2.6 = (Bo)^{0.5} << 1$. Therefore, we can assume $F_c$ to be a significant contributor in supporting the robot at the surface. The above expression also gives us $Bo$, Bond number, equal to $0.31$.

For simplicity, we consider the weight $F_W$, as shown in Fig.~\ref{stat_dyna}~(b), to act uniformly along the entire length of all the three legs. In other words, we assume uniform weight distribution on these legs. Let's, for a moment, assume the submerged angle $\phi$, as shown in Fig.~\ref{stat_dyna}~(a), is equal to $90^o$. In that case, the curvature force in vertical direction $F_c$ can be calculated as $F_c=2\sigma L \cos(\theta)$, where $L$ is the total length of the legs, and $\theta$ is the contact angle. From this expression, the maximum curvature force can be written as $2\sigma L$, which can be set equal to the weight to determine the minimum length of the legs, that turns out to be $L_{min}$=0.5~cm. However, this length of the legs is sufficient only when the entire supporting force due to surface tension acts in the vertical direction, which is not always the case. Also, the dimple due to a neighbouring leg can interfere with that of the one under study, which can reduce the supporting curvature forces as explained in \cite{song2007surface}. Additionally, ripples that are generated due to the ground effect of flapping wings have unknown effect on the legs, and it can also reduce the supporting force at the interface. Because of these unknown effects, we turn to empirical study conducted by Hu et al. in \cite{hu2003hydrodynamics}. A study of 342 species of water striders was conducted, and the relation between the maximum curvature force and body weight was found out. The plot of the relation is shown in Fig.~\ref{Hu_Bush_plot}. The best fit line of the plot of this data was given by $max(F_c) = 48 F_W^{0.58}$, where the forces are measured in dynes. Considering a $80$~mg robot, $max(F_c)\approx 600$~dynes, which corresponds to $L\approx4.2$~cm. This length of legs will add an extra mass of $\sim 15$~mg to the robot, which is compensated by further changing the total length of legs to $5$~cm. So, the set of horizontal legs now consists of three pieces with the center one $2.5$~cm long and other two $1.25$~cm each. RoboFly is also compared with other water striders in Fig.~\ref{Hu_Bush_plot}.

Special attention is given to the minimum distance between two adjacent legs. If they are too close, the dimple caused by a leg on the water surface will interfere with that by its neighbouring legs. This will reduce the lift force generated by the surface tension at the contact with legs, this is shown in the study by \cite{song2007surface} about the deformed water profile reducing the lift force. For a static case, it can be seen in their work that the dimple dies down at about $6$~mm from a floating object, which requires us to have at least $12$~mm of gap between legs. Keeping that in mind, we choose to have them $15$~mm apart to be in safe situation in dynamic case in which the robot will be performing water locomotion.

\subsection{Experimental results}

The signal generated in this case is similar to the one used for ground locomotion. When the wings are flapped, ripples can be seen generating and propagating away from the robot on the water surface. Though the effect of ripples is unknown on the motion of the robot, it is observed that the robot doesn't move in any direction when the two wings are driven by the same driving signal without any second harmonic component. It can be concluded that the ripples have equal effect in all directions, and thus can be ignored in the dynamic force balance in the horizontal plane.

\begin{figure*}[h]
    \centering
     \includegraphics[width=6.5in]{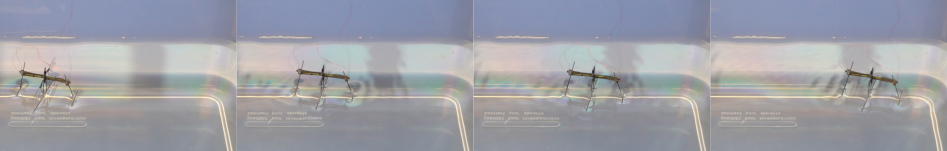}
     \caption{Robot performing water surface locomotion with open loop control by flapping wings at 35~Hz, frames are captured at 0, 4, 8, and 12~s}
     \label{water_walker_straight_line}
\end{figure*}

\begin{figure*}[h]
    \centering
     \includegraphics[width=6.5in]{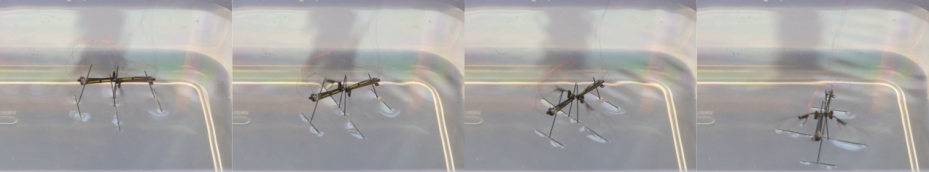}
     \caption{Robot performing water surface locomotion and turning left with open loop control by flapping wings at 30~Hz, frames are captured at 0, 1, 2, and 3~s}
     \label{water_walker_turn}
\end{figure*}

\begin{figure*}[h]
  \centering
  \includegraphics[width=6.5in]{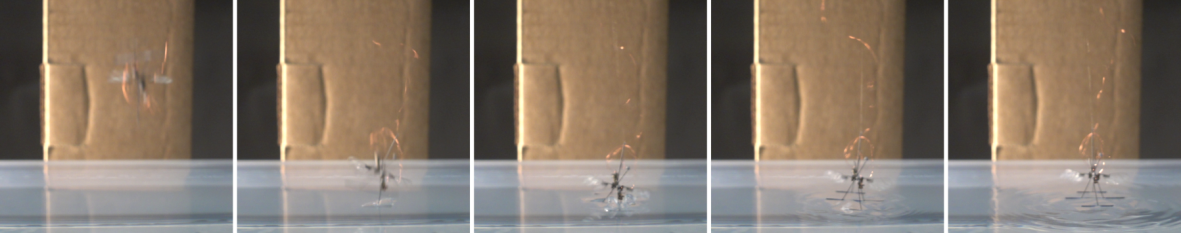}
  \caption{Airborne RoboFly landing on the water surface. Frames are captured at 0~ms (robot is still airborne), 30~ms (landing only on front leg), 45~ms (middle leg also gets in contact with the water surface), 95~ms (all three legs are in contact now) and 145~ms (oscillations are damped and the robot is stable).}
  \label{landing}
\end{figure*}

RoboFly with its horizontal legs is shown in Fig. \ref{Fly_water}. The design parameters for the legs are summarized in Table \ref{legs_table}. As mentioned in above section, RoboFly has the capability to move in either direction. Frames captured from a video of RoboFly moving from left to right are shown in Fig.~\ref{water_walker_straight_line}. The robot in this video was driven with a second harmonic signal of amplitude 220~V, and the wings were flapped at 35~Hz, far below its resonant frequency. The recorded speed of this interfacial flight was $\approx$0.5~cm/s. In addition to the straight-line motion, the robot can also steer by flapping one wing at a larger amplitude than the other. The robot can be seen taking a sharp turn towards left in Fig.~\ref{water_walker_turn} by flapping the right wing at a larger amplitude (220~V) than the left one (180~V). The angular speed is recorded as $\approx$20$^o$ per second. Fig. \ref{landing} shows a set of images in which an airborne robot is seen to be landing on the water surface. In this case, the robot simply jumps off the cardboard box in the background. It can be seen that the robot didn't land in its stable orientation; however, it still manages to recover without breaking the surface tension of the water. This can be attributed to the length of the legs being inspired by the actual biological species which may not have all the legs touching the water surface all the time. We were able to achieve this landing behavior consistently.

\begin{table}[!t]
\renewcommand{\arraystretch}{1.3}
\caption{Summary of design parameters of leg}
\label{legs_table}
\centering
\begin{tabular}{c c}
\hline
\hline
Design Parameter & Value/Characteristic/Name\\
\hline
Number of legs & 3 \\
Cross-section & Cylindrical \\
Material & Carbon fiber\\
Diameter & 0.5 mm\\
Total Length & 50 mm\\
Distance between adjacent legs & 15 mm\\
Hydrophobic Coating & \textit{Rust-Oleum NeverWet} \\
Contact angle & 135$^o$ (Fig.~\ref{contact_angle})\\
\hline
\end{tabular}
\end{table}

\subsection{Challenges}

When the robot tries to take off, it relies on the wing lift to generate enough thrust to break its contact with the water surface. Immediately preceding the contact breakage, the surface tension which pulls the robot back assumes its maximum value, $2\sigma L$. If we assume, in a failed attempt of take-off, the robot comes to a static position right before the contact breaks, the meniscus assumes the maximum height. For the static analysis at this position, we can set the buoyancy force $F_B$ equal to zero as robot legs are at their maximum height on the water surface. This will reduce the equation of motion to as in (\ref{eq:static_equation for liftoff}).
\begin{align}
\begin{aligned}
F_L =F_W+F_{C,max}
\end{aligned}\label{eq:static_equation for liftoff}
\end{align}

Now, substituting $F_{C,max}=2\sigma L$ and values of all the unknowns, we can see that the value of $F_L$ corresponds to $800$~mg of force, which is 10 times the weight of the robot. It is impossible for this insect scale robot to lift off with an additional load of 9 times its own weight. That means it has to rely on some external factor to break the contact with the water surface. This phenomenon can be attributed to the behavior that has been observed with its biological counterpart \textit{nymphaeae}. These insects, as shown in \cite{mukundarajan2016surface}, oscillates in the vertical direction as they generate lift only in the downstroke; therefore, the meniscus is observed to be assuming its maximum height during downstroke and maximum depth during upstroke. As \textit{nymphaeae} are seen to be taking off after some time which varies from a fraction of seconds to a few seconds, once the oscillations assume large amplitudes and the inertia is large enough to break the surface. 

In \textit{nymphaeae} study, the required lift-to-weight ratio is observed to be $q=3.4$, whereas it is observed to be $q\approx10$ in the case of its robotic counterpart. The transition from iterfacial flight to airborne flight with the help of induced oscillations is considered to be out of scope for this paper.

\subsection{Comparison with other biological locomotion}

The locomotion of aquatic and semi-aquatic insects can be characterized using the thrust generating mechanism. The water surface locomotion shown by RoboFly is similar to most semiaquatic insects which utilize their hydrophobic legs for flotation and propulsion. This locomotion is known as water walking. Water walking insects also rely on surface tension for flotation as reviewed in \cite{hu2003hydrodynamics,hu2010hydrodynamics,bush2006walking}. 

Another locomotion inspired by Marangoni propulsion of rove beetle and semiaquatic insects like \textit{Microvelia} and \textit{Velia} is reviewed in \cite{bush2006walking}. In this locomotion, the thrust is generated by uneven surface tension because of a chemical released by the insects.

Honeybees, when trapped on the water surface, show a unique type of locomotion referred to as hydrofoiling locomotion by \cite{roh2019honeybees}. Honeybees are seen to be using their wings as hydrofoils to generate hydrodynamic thrust. Another interesting locomotion on water surface similar to honeybee's hydrofoiling locomotion is the rowing locomotion shown by the stonefly as reported in \cite{marden2003rowing}. This locomotion makes use of both hydrodynamic drag as well as aerodynamic lift for propulsion. 


\section{Power Consumption}

The cost of transport (CoT) is a useful metric to compare different modes of locomotion. CoT is defined as the energy expense per unit distance traveled, or equivalently the power required per unit velocity, as shown in Eq. \ref{eq:performanceMetricEqn}. To calculate CoT for the robot, voltage and current supplied to the bimorph actuators was measured at a sampling frequency of $10$~kHz, from the voltage and current monitor outputs of the amplifiers (Trek 2205,
Lockport, New York). The instantaneous power was time averaged over an integer number of wing strokes in order to compute the average power consumption of the robot. 

In order to estimate power requirements for an untethered robot, we envision a linear half-bridge  driver circuit such as demonstrated in \cite{JJames_ICRA2018}. In comparison to the desktop amplifiers, the onboard linear actuator driver would add inefficiency which would increase power requirements above what we measure for the tethered robot powered by the desktop amplifiers. Therefore, to estimate the power requirements an untethered robot, reverse power from the center node of the parallel-connected bimorph actuator is zeroed when computing the integral of $V, I_r$ in Eq. \ref{eq:performanceMetricEqn}. This assumption reflects the reality for linear half-bridge piezo driver methods that during the part of the wing stroke in which the sinusoidal drive signal voltage is decreasing, positive charge leaving the center node of the actuator must be dumped to ground through the transistor elements of the driver, and that energy cannot be recovered to the high voltage bias rail or back to the boost converter power source. Alternate driver topologies which are capable of bidirectional power flow \cite{jafferis2019untethered} or which implement the energy recovery discussed in \cite{Karpelson2012}, can be somewhat more efficient than this proposed linear driver, so the power consumption and $CoT$ computed here provides an upper bound of power requirements. The integral of the measured voltage $V$ and current $I$ from the amplifiers represents the lower bound of predicted power autonomous robot requirements because that would assume perfect driver efficiency.  

Eq. \ref{eq:performanceMetricEqn} shows the computation of the cost of transport ($CoT$), that is, the energy used per unit distance traveled. The measured power was integrated over time and divided by the estimated distance traveled by the robot as measured by the motion capture system.
\begin{equation} \label{eq:performanceMetricEqn}
CoT = \dfrac{\int  VI_{r} ~\mathrm{d}t }{S}
\end{equation}

Where $S$ is the 3D distance traveled, $V$ is the driving voltage, and $I_r$ is the measured current. 

\begin{figure}[tb]
	\centering
	\includegraphics[width=3.2in]{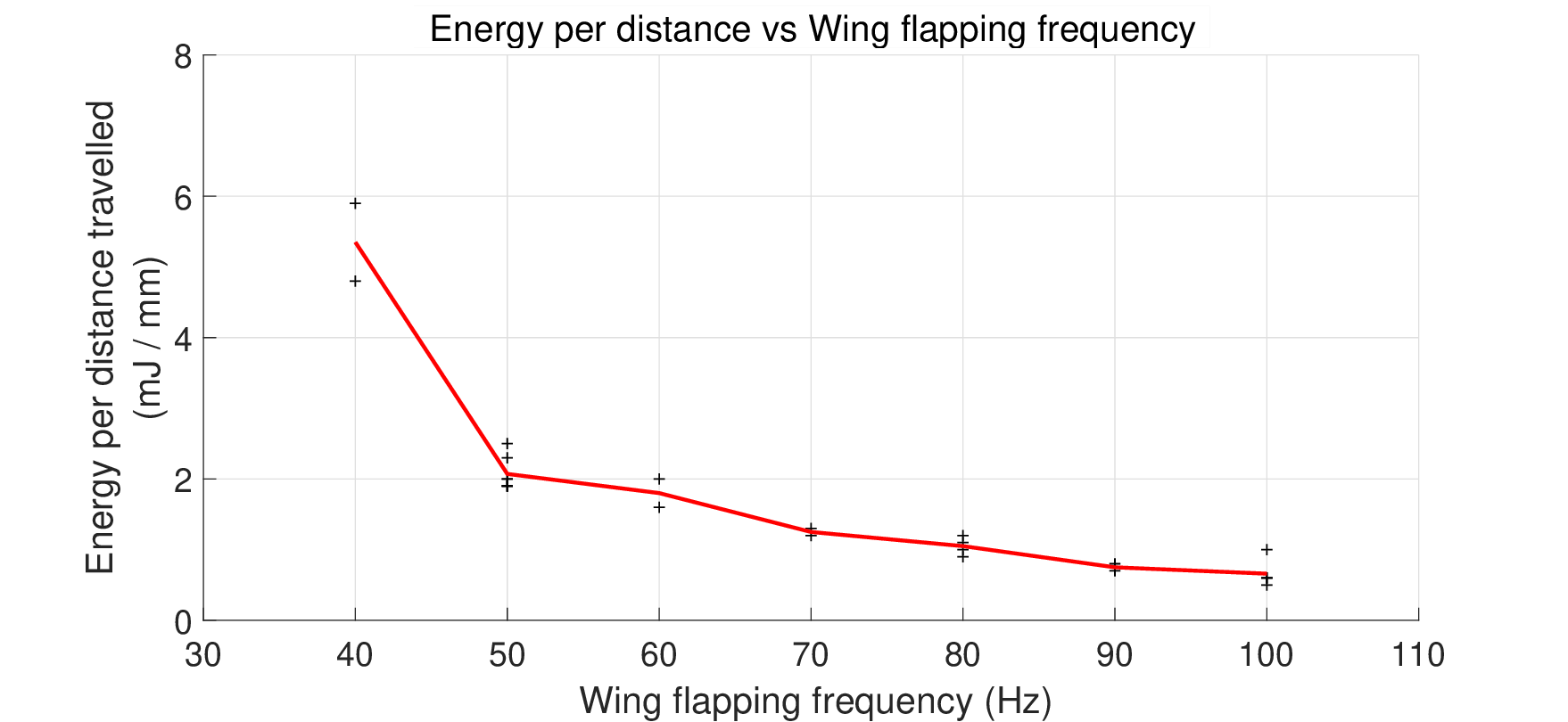}
	\caption{The cost of transport ($COT$), the energy expended per unit distance traveled, decreases with increasing drive frequency $f$. }
	\label{power_image}
\end{figure}

To measure CoT for ground ambulation, we set the driving amplitude to $250$~V. The results show that the cost of transport decreases with increasing flapping frequency (Fig.~\ref{power_image}). This is conjectured to be the result of two competing factors: 1) Electrical input power increases proportionally with driving frequency $f$ because the actuators are principally a capacitive load with current $ I \propto C\frac{\mathrm{d}V}{\mathrm{d}t}$; 2) The aerodynamic lift increases with $f^2$ for the same reason that drag does (Section III). Therefore, for small frequency increases the robot is expected to reduce contact friction during ambulatory and water-striding motion faster than the power requirements increase. Together these factors suggest an inverse proportionality between $CoT$ and flapping frequency which is observed in Fig. \ref{power_image}.

Measurements also indicated the trend of decreasing $COT$ with increasing frequency while the robot was in flight. The $COT$ was measured in flights traversing $0.2$~m as measured by motion capture while flapping at $140$~Hz. The $COT$ for flying locomotion was $\sim 0.02$~mJ/mm, which is $\approx 25 \times$ less than the most efficient ambulation. $CoT$ is not an applicable metric for hovering flight at a single position. The robot consumed approximately $50$~mW measured power during hovering flight with feedback controlled attitude and position. 

\section{Future Work}
Water locomotion can be further improved by performing a transition from water surface to airborne flight. We found that the vehicle was unable to lift off from the surface if the water simply by flapping its wings, as predicted by our earlier analysis indicating that the robot needs to lift approximately 9 times its own weight to break the water surface tension. In addition to explosive ejection \cite{chen2017biologically}, other potential means include: 1) electrowetting of the legs to reduce the contact angle and therefore reduce the force \cite{chen2018controllable}, 2) creating waves on water surface with the help of flapping wings to lower the force required to break water surface film, 3) making the robot collide with an obstacle on the water surface which could provide the necessary impulse force.

As to ground locomotion, alternative modes such as jumping are possible. Hopping locomotion can be efficient due to advantageous scaling effects as robot size and weight are reduced \cite{PrinciplesOfAnimalLocomotion}, although additional weight and complicated hopping mechanisms are ill suited to honeybee-sized flying robots as discussed in Section III. Fei in \cite{FeiLi2012} demonstrates bio-inspired jumping mechanisms and discusses the dynamics and optimization. Bhushan and Tomlin in \cite{bhushan2019insect} demonstrates an insect-sized microrobot capable of jumping 6 times per minute.

This work uses a 2-component Fourier basis to cause differential stroke speed, but we have not explored other waveforms e.g. `sawtooth', which could perform better.

Although inertial dynamics of the flapping motion were experimentally determined to be less significant than the aerodynamic drag acting on the wings (Section \RNum{2}), there is still uncertainty as to the role of vibratory mechanisms in the ground locomotion. If vibratory mechanism is significant, then the feet could be redesigned to exploit this; e.g. directional spines could serve to selectively favor a direction of motion.

\section{Conclusion}

This paper presents a new design with three major contributions to the field of insect-sized robotics. It 1)~simplifies fabrication, 2)~allows the robot to perform landing and ground locomotion, 3)~designs legs for the robot to perform water surface locomotion similar to its biological counterparts.

In the new design, the airframe and transmission are all folded from a single laminate sheet. Compared to previous work, the design presented here represents an intermediate solution that lies between the many parts of \cite{ma2013,ma2012} and the single laminate sheet composed of many layers of \cite{Sreetharan2012} (Table~\ref{table_comparison}). We believe this represents a valuable intermediate between these two extremes because on the one hand our design with two laminates gains many of the benefits of pop-up book manufacturing, such as having few parts and the ability to precision align small components. And on the other hand, it does not inherit the substantial complexity imposed by large number of interdependencies among layers. This reduces the difficulty of design iteration. Furthermore, we believe our intermediate approach is still amenable to automated manufacturing, by assuming that some steps will be performed by small robotic end-effectors. 

We showed that the lowered center of gravity of the robot allows it to land and ambulate along the ground including steering, in addition to flight. It was able to land consistently under feedback control and it was even able to land upright from an unstable open-loop flight. The cost of transport was found to be substantially higher than that of free-flight, so this mode of locomotion is better suited to precise motions, such as to precisely position a sensor. We additionally showed that ground ambulation can allow our robot to reach new places that are not accessible through flight, such as moving under a typical door. This represents a capability to negotiate an obstacle that heretofore exclusively the domain of the most adept ground robots, and impossible with air robots. 

We also demonstrated another locomotive capability of the robot where it can, with the help of a set of three small horizontal legs, land on water surface and perform a waterlily beetle-like locomotion along the surface. 

Our robot's multi-modal locomotion capabilities resemble those of larger robots. For example, \cite{daler} developed a larger bio-inspired robot (393~g, 72~cm) capable gliding flight as well as the ability to ambulate by rotating its ailerons. \cite{bachmann} developed a bio-inspired micro-vehicle (100~g, 30.5~cm) capable of performing aerial locomotion using wings and terrestrial locomotion using whegs. Similarly, \cite{peterson} developed a bipedal ornithopter (11.4~g, 28~cm) with flapping wings for aerial locomotion and rotary legs for terrestrial locomotion. A 30~g robot took an approach similar to our robot by using the four propellers of its flight apparatus to steer its motion. These were used to steer a simple walking mechanism that was capable of moving in only one direction \cite{Fly_monkey}. To our knowledge this work represents the first example of multi-modal locomotion capability at insect scale. 

The capability of landing will allow the robot to perform intermittent flights. This will be useful for providing power to the robot. For example, the robot could more easily collect power from a laser because the laser would not have to follow it~\cite{JJames_ICRA2018, LaserPhoneChrgr, LaserMotive}, or from magnetic resonance coupling, as has previously been demonstrated on a ground robot in~\cite{Karpelson2014}. Furthermore, landing will be necessary for the robot to collect energy from ambient energy sources such as indoor light or radio frequency signals such as WiFi \cite{PowWifi} or cellular. In the case of energy harvesting from aeroelastic flutter \cite{flutterHarvest}, ground locomotion may be needed to position the robot in the flow. While these sources tend to be very minute and therefore insufficient to power larger robots, they may be enough to power the UW RoboFly for a reasonable fraction of the time, if it can land and charge between flights. The horizontal design of this work facilitates the attachment of power electronics \cite{JJames_ICRA2018} and sensors such as ultralight cameras \cite{siva_biorob18}.

\section*{Acknowledgment}

This work is partially supported by the Air Force Office of Scientific Research under grant no. FA9550-14-1-0398. The authors also wish to thank Haripriya Mukundarajan, PhD student at Prakash Lab, Stanford University for insightful discussions during the study on water surface locomotion.

\bibliographystyle{IEEEtran}
\bibliography{IEEEabrv,citations}

\begin{IEEEbiography}[{\includegraphics[width=1in,height=1.25in,clip,keepaspectratio]{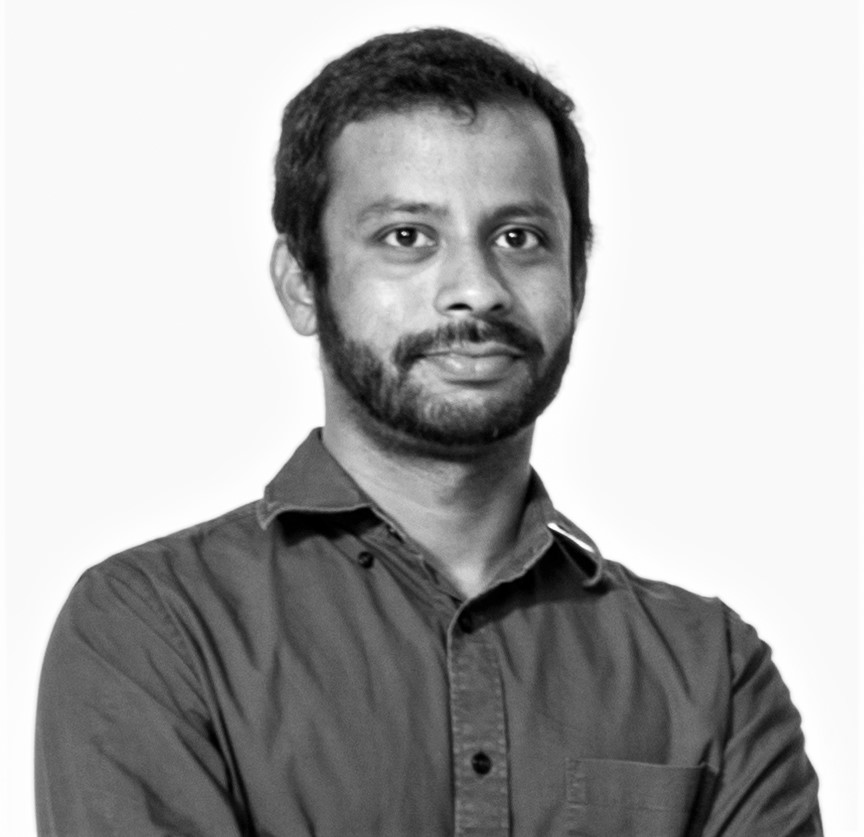}}]%
{Yogesh Chukewad}
received the Ph.D. in mechanical engineering from the University of Washington, Seattle, WA. in 2020.

His research interests include design and controls of micro-robots, sensor fusion and localization for autonomous robots. 
\end{IEEEbiography}

\begin{IEEEbiography}[{\includegraphics[width=1in,height=1.25in,clip,keepaspectratio]{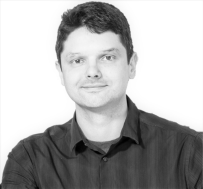}}]%
{Johannes James}
received the B.S. in mechanical engineering and USCG merchant mariner engineering license from the California Maritime Academy C.S.U. in Vallejo, California, U.S.A. in 2013. He is currently working toward power autonomous flapping wing insect robots pursuant to the Ph.D. degree in the Mechanical Engineering dept. at the University of Washington. 

His research interests include industrial automation and control, embedded systems, micro-robotics, and power electronics.
\end{IEEEbiography}

\begin{IEEEbiography}[{\includegraphics[width=1in,height=1.25in,clip,keepaspectratio]{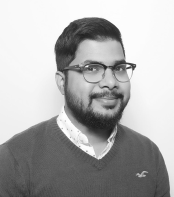}}]%
{Avinash Singh}
received his B.E. in mechanical engineering from Nagpur University, India in 2013 and his M.S. in mechanical engineering from the University of Washington, Seattle, WA, USA in 2018. He is currently working as a robotics engineer in the Seattle, WA.

His research interests include autonomous robotics, bio-inspired robotics, aerial robotics, SLAM and machine learning.
\end{IEEEbiography}

\begin{IEEEbiography}[{\includegraphics[width=1in,height=1.25in,clip,keepaspectratio]{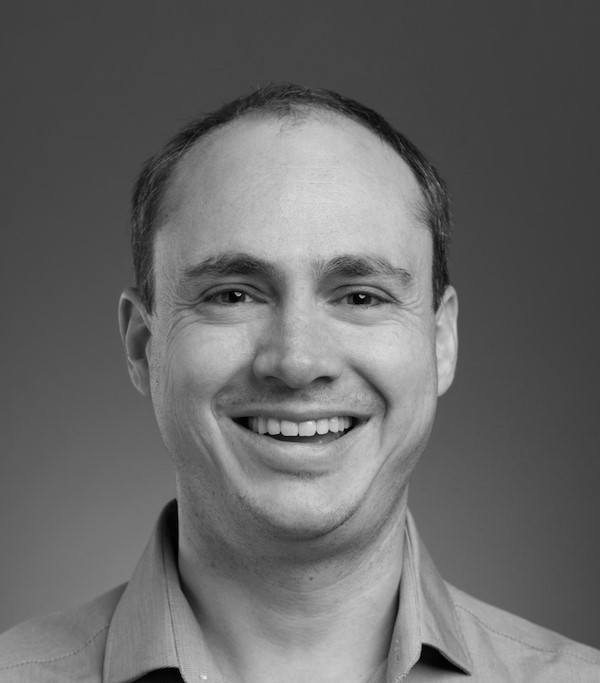}}]%
{Sawyer Fuller}, Assistant Professor of Mechanical Engineering, creates biologically-inspired sensors, control systems, and mechanical designs targeted at insect-sized air and ground vehicles, and investigates the flight systems of aerial insects. He completed his Ph.D. in Biological Engineering at the California Institute of Technology and B.S. and M.S. degrees in Mechanical Engineering at the Massachusetts Institute of Technology, and postdoctoral training at Harvard. In addition to his work in insect flight control, he also developed a frog-hopping robot at the NASA Jet Propulsion Laboratory and invented an ink-jet printer capable of fabricating millimeter-scale 3D metal machines at the MIT Media Lab. 
\end{IEEEbiography}

\end{document}